\pdfoutput=1
\documentclass{article}

\PassOptionsToPackage{numbers, compress}{natbib}

\usepackage[final,dblblindworkshop]{neurips_2026}
\workshoptitle{Foundation Models for Temporal Systems: From Forecasting to World Modeling}

\usepackage[utf8]{inputenc}
\usepackage[T1]{fontenc}
\usepackage{hyperref}
\usepackage{url}
\usepackage{booktabs}
\usepackage{graphicx}
\usepackage{placeins}
\usepackage{amsmath}
\usepackage{amsfonts}
\usepackage{nicefrac}
\usepackage{microtype}
\usepackage{xcolor}

\title{KITE: A Knowledge-Injected Temporal Encoder for Time-Series Event Forecasting}

\author{%
  Jonas Petersen$^{1}$\quad
  Pratik Jawahar$^{2}$\quad
  Riccardo Maggioni$^{1}$\quad
  Gian-Alessandro Lombardi$^{2}$\\
  \textbf{Matteo Cozzi$^{1}$\quad
  Marek Gajewski$^{2}$\quad
  Federico Martelli$^{2}$\quad
  Camilla Mazzoleni$^{1}$}\\
  \textbf{Philipp Christian Petersen$^{3}$}\\[0.6em]
  $^{1}$ETH Zurich\quad
  $^{2}$Forgis AG\quad
  $^{3}$University of Vienna\quad
  \texttt{jep79@cantab.ac.uk}
}

\hypersetup{
  pdftitle={KITE: A Knowledge-Injected Temporal Encoder for Time-Series Event Forecasting},
  pdfauthor={Jonas Petersen, Pratik Jawahar, Riccardo Maggioni, Gian-Alessandro Lombardi, Matteo Cozzi, Marek Gajewski, Federico Martelli, Camilla Mazzoleni, Philipp Christian Petersen}
}

\begin{document}

\maketitle

\begin{abstract}
  Forecasting rare, high-cost events in multivariate time-series systems is hard because their precursors are seldom observed, so a purely data-driven model has little to learn from. The engineers who run these systems, however, hold knowledge that precedes such events, information absent from the scarce labels. We call this a \emph{semantic prior} and inject it into a self-supervised event forecaster through two switchable mechanisms, a covariate mechanism and an attention-bias mechanism, each the identity at its null, so one architecture serves every prior type with no per-dataset architecture search. A prior helps when it supplies information the observed window does not already contain: in a short context an informative prior beats no prior in-domain (pooled h-AUROC over the no-prior baseline: a periodicity clock $+14\%$, $p<0.001$; a conservation residual $+47\%$ at a $0.25$ label fraction), and a working prior transfers out-of-domain ($+53\%$), substituting for feature transfer, while priors the window already recovers stay inert and the attention-bias mechanism shows no measurable gain. KITE is a switchable, attributable route to contextual conditioning, and a substrate for scaling toward cross-domain, pretrained time-series foundation \emph{world models} that carry their governing structure explicitly.
\end{abstract}

\section{Introduction}

Multivariate time-series systems, building-management arrays, water networks, air-quality monitors, power infrastructure, are watched by dense arrays of timestamped variables, and the events that matter most are rare and expensive to label, so a purely data-driven model has little to learn from on exactly the stratum that matters. A domain engineer, by contrast, holds knowledge that precedes any data, which variable is the high-voltage load, that a reading above a threshold is a warning, that the daily cycle repeats every $1440$ minutes, that a pump drives a downstream temperature. This a-priori knowledge, which we term a \emph{semantic prior}, is exactly the information the scarce labels do not carry.

We use \emph{semantic prior} for engineering or domain knowledge, available before any data is seen, about the variables of a system and their relations: a variable's \emph{identity}, a \emph{bound}, a \emph{periodicity}, a governing \emph{conservation} law, or a \emph{causal edge}. That domain knowledge can act as a prior is an old idea, from classical feature engineering to Bayesian priors and inductive bias~\citep{vonrueden_informed_2023}; the question we take up is how to inject such heterogeneous knowledge into a self-supervised event forecaster in a way that is general across prior types, simple to apply, and honest about attribution.

Existing approaches address only slices of this problem: each injects a single prior type through an always-on pathway entangled with the backbone (Section~\ref{sec:related}), so a new prior or a new system means a new architecture, and none unifies the covariate and relational routes across the range of priors, bounds, conservation laws, periodicity, that engineers rely on.

KITE (Knowledge-Injected Temporal Encoder) addresses this gap by treating a prior as a \emph{switchable input} to a single self-supervised backbone rather than a change to its architecture. Built on a joint-embedding predictive architecture (JEPA) for multivariate time series~\citep{assran_self-supervised_2023,petersen_hepa_2026}, it routes prior types through two mechanisms: mechanism~(I), a \emph{covariate mechanism}, admits per-variable and derived priors as synthetic variables (a static descriptor to the encoder, a future-known signal to the horizon-conditioned predictor, a past-only derived law such as a conservation residual to a covariate); mechanism~(II), an \emph{attention-bias mechanism}, biases encoder attention with a relational prior such as a causal graph. Relative to the JEPA backbone, only the future-known covariate path and the whole of mechanism~(II) are new; the rest is routing. Because a prior enters as an input, one architecture serves every prior type with no per-dataset architecture search, and each prior-versus-no-prior comparison is a clean ablation. The prior's weights are still learned: a static or past-known prior enters Stage-1 pretraining, so a new one needs a new Stage-1 run, while a future-known prior enters only the predictor at Stage~2. Our contributions are: (1) a taxonomy of twelve semantic-prior categories mapped to these two mechanisms, of which four are implemented and evaluated here and the rest chart the design space; (2) the KITE architecture, built on a permutation-invariant, cardinality-free encoder so one model runs across systems with different variables; and (3) an empirical study organised by a single principle, \emph{a prior helps when it supplies information the observed window does not already contain}: in a short context an informative prior beats no prior in-domain (Occupancy clock $+0.104$, $p<0.001$; BATADAL conservation residual $+0.741$) and transfers out-of-domain (BATADAL$\to$SYNTH-CON $+0.32$) by substituting for feature transfer, and an account of where priors are inert (redundant static injection, the graph bias, long context).

\section{Related work}
\label{sec:related}

The dominant context for KITE is the wave of pretrained time-series foundation models, from Chronos-2~\citep{ansari_chronos-2_2025}, TimesFM~\citep{das_decoder-only_2024}, and Moirai, whose any-variate attention accepts arbitrary variable counts~\citep{woo_unified_2024}, to the variate-token designs PatchTST~\citep{nie_time_2023} and iTransformer~\citep{liu_itransformer_2024}, the growing push to pair series with structured and textual context~\citep{liu_rethinking_2026}, and industrial pretraining corpora that map heterogeneous machines onto a shared variable schema~\citep{othman_factorynet_2026}: these transfer with little retraining but forecast future \emph{values} and expose no switchable pathway for a-priori domain knowledge. Injecting a prior is itself not new, but each line handles one type through an always-on pathway trained into the weights: covariate-conditioned forecasters~\citep{lim_temporal_2020,han_unica_2026,yamaguchi_citras-fm_2026}, graph-conditioned models~\citep{zhao_multivariate_2020,deng_graph_2021,li_diffusion_2018,yu_spatio-temporal_2018}, physics-informed losses~\citep{raissi_physics-informed_2019}, architecture-baked periodicity~\citep{wu_autoformer_2021,wu_timesnet_2023}, and text-conditioned JEPA encoders that treat descriptions as always-on identifiers~\citep{dutta_giving_2026}. KITE builds on the JEPA family~\citep{assran_self-supervised_2023,petersen_hepa_2026} and unifies covariate, relational, conservation, and periodicity priors as switchable inputs to one backbone, each the identity at its null, so a prior's contribution is separable from a pathway's capacity.

\section{Method}

\begin{figure*}[t]
  \centering
  \includegraphics[width=\textwidth]{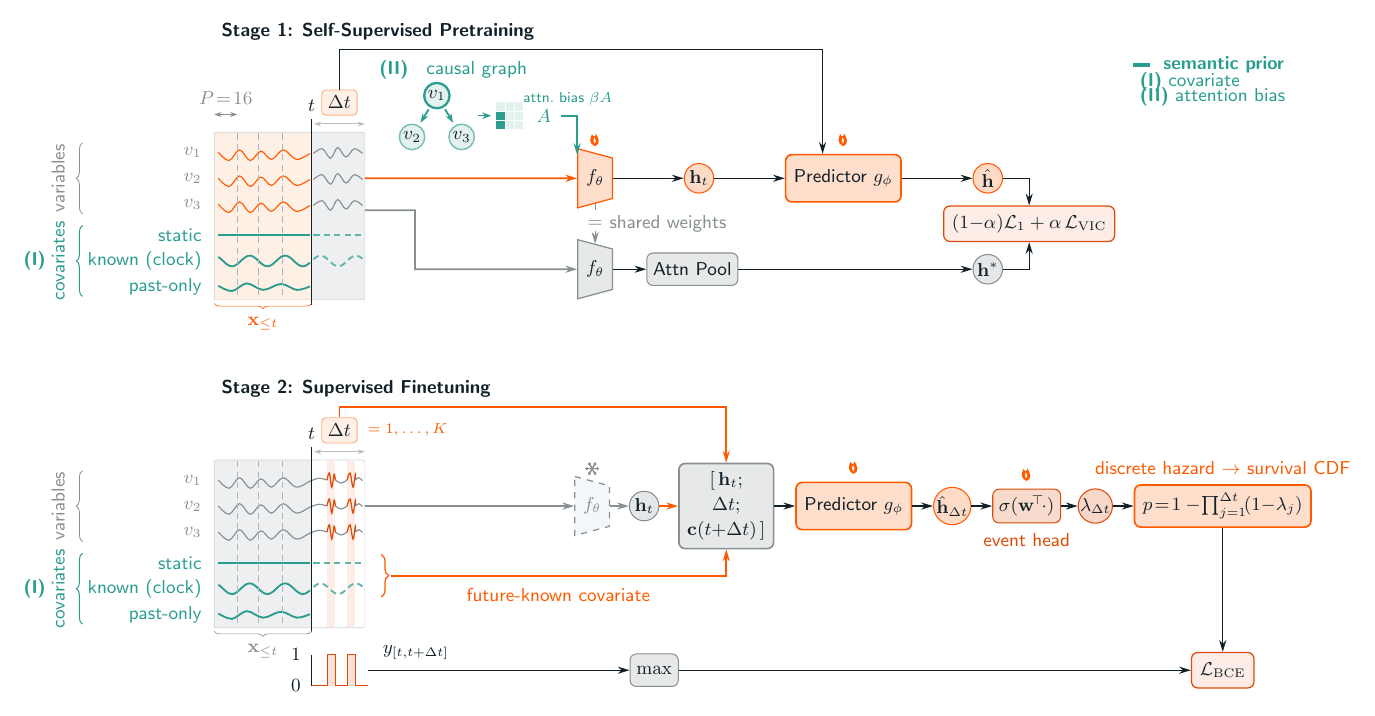}
  \caption{Two-stage KITE architecture. \textbf{Stage 1 (pretraining):} a funnel encoder with causal-graph attention bias ingests the context; static/past-known covariates enter the encoder while future-known covariates enter the horizon-conditioned predictor, and a shared-weight target encoder (not EMA) supplies the latent-prediction target. The causal-graph bias acts only in Stage-1 pretraining (the encoder is frozen at Stage~2); its graph is \emph{per-dataset}, and in the reported runs its gate stays at the null $\beta=0$ (Section~\ref{sec:mechanisms}, Appendix~\ref{app:graph}). \textbf{Stage 2 (finetuning):} a compact head over the frozen encoder consumes a past/future by variable-type input grid and emits a monotone event-probability surface $p(t,\Delta t)$.}
  \label{fig:architecture}
\end{figure*}

\subsection{The semantic-prior taxonomy}
\label{sec:taxonomy}

The organising idea is that domain knowledge can be sorted by \emph{when it is knowable} and \emph{whether it is per-variable or relational}, which determines where the prior enters. The twelve categories are listed in Appendix~\ref{app:taxonomy}; we implement and evaluate four, periodicity (\#5), conservation (\#10), a structural descriptor (\#1), and causal-graph attention bias (\#9). Static per-variable priors were tested and found inert, motivating the routing view over static injection.

\subsection{Backbone: encoder, predictor, and event head}
\label{sec:backbone}

KITE's contribution is the semantic-prior conditioning; the backbone is a horizon-conditioned event JEPA~\citep{petersen_hepa_2026}. A variate-token encoder $f_\theta$ maps a window to a summary $\mathbf{h}_t$, patchified into $(C\times N)$ tokens with \emph{no per-variable slot} (RoPE-ordered patches, variables an unordered set), so it is permutation-invariant and cardinality-free ($C_{\text{test}}\ne C_{\text{train}}$; Appendix~\ref{app:perm}). A horizon-conditioned predictor $g_\phi(\mathbf{h}_t,\Delta t)$ and a discrete-survival head compose per-horizon hazards $\lambda_k$~\citep{gensheimer_scalable_2019,kvamme_continuous_2021} into a monotone event-probability surface $p(t,\Delta t)=1-\prod_{j\le k}(1-\lambda_j)$. Stage~1 pretrains encoder and predictor self-supervised (shared-weight target, $L_1$ latent loss with a VICReg~\citep{vicreg} collapse term); Stage~2 freezes the encoder and trains the head on labels. We evaluate by the rank-based horizon-AUROC (h-AUROC).

\subsection{Semantic-prior mechanisms}
\label{sec:mechanisms}

KITE injects knowledge through two mechanisms, each the identity at its null, so at initialization the conditioned model computes exactly the unconditioned one (Figure~\ref{fig:architecture}; full per-path detail, equations, and the graph-bias schematic are in Appendix~\ref{app:mech}). \textbf{Mechanism~(I), covariates}, routes per-variable and derived priors as synthetic variables along three paths, by \emph{when} the signal is knowable: a \emph{static} descriptor to the encoder (a zero-initialized fusion $\mathrm{out}=x+W_\delta\delta$, present in pretraining), a \emph{future-known} covariate to the horizon-conditioned predictor (a periodicity clock evaluated at $t{+}\Delta t$, so it anticipates rather than detects), and a \emph{past-only derived} law to a covariate (a conservation residual $\sum_{in}-\sum_{out}-dL/dt$, feature engineering over an adaptable model). \textbf{Mechanism~(II), attention bias}, adds a scalar-gated driver-adjacency $\beta A$ to the encoder's attention logits, identity at $\beta{=}0$; it does not measurably help here (bounded to $\approx 0.006$ h-AUROC, Appendix~\ref{app:graph}).

\section{Results and Discussion}
\label{sec:results}

We test this principle, that a prior helps when it supplies information the window lacks, in two regimes, in-domain short context (Claim~A) and out-of-domain transfer (Claim~B), by h-AUROC over $5$ to $10$ seeds against zeroed- and shuffled-prior controls, so a gain cannot be an artefact of the extra input. Identity at the null holds only at fixed weights, and training with a prior also adapts the rest of the network, so every reported null is a separately trained model with the prior zeroed or absent, never the prior model with its input removed at test time.

\begin{table}[t]
  \centering
  \caption{Semantic priors against external baselines (h-AUROC, mean$\pm$sd over $5$ seeds unless noted; higher is better; ``det.''\ = training-free deterministic detector). \textbf{No prior} is the prior-free control under the same recipe (zeroed on the BATADAL rows, absent elsewhere; the two agree on Occupancy, $0.767$ vs.\ $0.762$); the external forecasters have no semantic-prior pathway. \textbf{The estimand differs by row and is comparable only within a row} (stratum-restricted h-AUROC on the onset/ongoing rows, overall on the $0.25$-label row), so \emph{No prior} values below $0.5$ are not errors: on the hard stratum the prior-free backbone anti-ranks the rare-event windows. In the transfer row, \textbf{N/A} marks supervised baselines, which have no transfer mode; Chronos-2 is scored zero-shot on the same $2507$ windows ($225$ onset), window-bootstrap $95\%$ CI $[0.718,0.822]$. The two KITE transfer cells share one Stage-1 encoder; the prior arm scores $0.922$ whether it was pretrained on the target, the source, or not at all.}
  \label{tab:main}
  \footnotesize
  \setlength{\tabcolsep}{4pt}
  \resizebox{\textwidth}{!}{%
    \begin{tabular}{@{}llccccc@{}}
      \toprule
      Dataset (estimand)             & Prior (\#)                       & KITE (prior)   & No prior & Chronos-2         & PatchTST          & iTransformer      \\
      \midrule
      \multicolumn{7}{@{}l}{\emph{Short-context, in-domain (Claim A)}}                                                                                          \\
      Occupancy (silent-onset)       & Periodicity clock (\#5)          & \textbf{0.866} & 0.762    & $0.733{\pm}0.003$ & $0.702{\pm}0.037$ & $0.754{\pm}0.025$ \\
      BATADAL (ongoing)              & Conservation resid.\ (\#10)      & \textbf{0.877} & 0.136    & $0.947$ (det.)    & $0.732{\pm}0.073$ & $0.672{\pm}0.055$ \\
      BATADAL (overall, $0.25$ lab.) & Conservation (\#10)              & $0.644$        & 0.437    & $0.642$ (det.)    & $0.656{\pm}0.024$ & $0.534{\pm}0.036$ \\
      SYNTH-ROUTE-v2 (onset)         & Structural descr.\ (\#1, synth.) & $0.892$        & 0.674    & $0.909{\pm}0.007$ & $0.710{\pm}0.065$ & $0.916{\pm}0.011$ \\
      \midrule
      \multicolumn{7}{@{}l}{\emph{Out-of-domain transfer (Claim B)}}                                                                                            \\
      BATADAL$\to$SYNTH-CON (onset)  & Conservation resid.\ (\#10, OOD) & \textbf{0.922} & 0.601    & $0.774$           & N/A               & N/A               \\
      \bottomrule
    \end{tabular}%
  }
\end{table}

\subsection{Claim A: an informative prior beats no prior in short context}

Table~\ref{tab:main} carries Claim~A (bars: Figure~\ref{fig:results}). On the flagship Occupancy silent-onset stratum under a short $w=360$ context, the true-phase periodicity clock lifts the prior-free baseline from $0.762$ to $0.866$; with the clock on both paths, a zeroed clock stays at baseline ($0.767$) and a shuffled one falls $0.346$ below the true phase, so the gain is the phase itself. Where the clock enters matters: alone, as a time-of-day sine, it scores $0.721$; as ordinary encoder input channels, the way a generic covariate-capable forecaster takes it, $0.797$; only at the predictor's horizon $0.866$ (Appendix~\ref{app:prioronly}). On BATADAL at $w=512$ the prior-free null sits below chance, and the conservation residual lifts it to $0.877$ ($p=1.2\times10^{-5}$, $n=5$), still helping at a $0.25$ label fraction ($+0.207$). A synthetic structural descriptor (\#1) confirms the static path works when a genuine structural prior exists ($p=0.030$).

Against the external baselines of Table~\ref{tab:main}, none of $20$ arms beats KITE at BH-FDR $0.05$; they sit at or below No prior on Occupancy and well above it elsewhere, some with point estimates above KITE's. Our claim is that the prior lifts a backbone above its own null at no extra training cost, not that KITE beats foundation models.

\paragraph{Claim B: transfer by substitution, not amplification.} A domain-invariant prior (a calendar clock, a mass-balance residual) survives a domain shift that scrambles learned features: across ten cells it lifts the forecasting stratum by $+0.12$ to $+0.39$ where it applies, while a static per-variable prior does not ($+0.006$). The sharper prediction that it should help \emph{more} out of domain than in does not survive (OOD$-$ID $=-0.002$); instead the data show \emph{substitution} (Appendix~\ref{app:transfer}): a prior with no pretraining beats transfer with no prior in eight of ten cells, and zero-shot Chronos-2 ($0.774$) sits between KITE with the prior ($0.922$) and without it ($0.601$).

\subsection{Where priors do not help}
\label{sec:negative}

The boundaries are as informative as the wins (Appendix~\ref{app:alignment}): a prior is inert when the window already carries the information (static injection $0/19$ under BH-FDR; the clock null once the context spans a full period; the graph bias bounded to $\approx 0.006$), when it is misaligned (a wrong period reverses the Occupancy win, $-0.022$), and when a present-state signal is asked to anticipate (the residual detects but does not forecast onset, and on the ongoing stratum a zero-parameter threshold on it alone, $0.921$, outranks KITE: that win is the residual's; Appendix~\ref{app:prioronly}); the pre-registered Weather control passes as a falsification check ($\Delta=-0.017$, $p=0.69$). The evidence is scoped to short context and two priors, several taxonomy categories remain conceptual, mechanism~(II) is bounded but undetermined, and the strongest transfer target is synthetic; Appendix~\ref{app:implalign} verifies the shipped configuration.

\section{Conclusion}

KITE injects domain knowledge into a self-supervised event forecaster as a switchable input. A prior helps when it supplies information the observed window lacks: in short context it beats no prior, and out of domain it substitutes for feature transfer, while redundant static injection and the graph bias do not. Never entangling the backbone, a switchable prior is a natural substrate for pretrained, cross-domain time-series foundation \emph{world models} that carry their governing structure explicitly.

\bibliographystyle{unsrtnat}
\bibliography{kite}

\appendix

\section{Mechanism details}
\label{app:mech}

Each mechanism is the identity at its null: with the prior input zeroed the pathway adds nothing, so at initialization the conditioned model is exactly the unconditioned one. After training the remaining weights have adapted to the prior, so the no-prior comparisons of Section~\ref{sec:results} use separately trained arms rather than removing the prior at test time.

\paragraph{Static path (encoder).} A static per-variable descriptor (categories \#1--\#4, \#6--\#8, \#12) is fused through a pathway that is identity at initialization,
\begin{equation}
  \mathrm{out} = x + W_\delta\,\delta, \quad W_\delta \text{ initialized to } 0 ,
  \label{eq:gated-pathway}
\end{equation}
present during Stage-1 pretraining so the frozen encoder learns to use it. Identity comes from the \emph{weights}, not a scalar gate (a $\tanh(g)$ gate opens slowly and costs much of every flagship effect, Appendix~\ref{app:implalign}). Generic descriptors are inert where the window already carries what they encode; the gain appears on a synthetic benchmark built to \emph{hide} identity.

\paragraph{Future-known path (predictor).} A future-known prior enters as $c(t+\Delta t)\in\mathbb{R}^m$ concatenated into the predictor input $[\mathbf{h}_t;\Delta t;c(t+\Delta t)]$ and evaluated at the horizon, so it anticipates rather than detects. The tested instance is a periodicity clock,
\begin{equation}
  c(t+\Delta t) = [\sin, \cos]\!\left(2\pi \, \frac{(t+\Delta t) \bmod P}{P}\right) ,
  \label{eq:periodicity-clock}
\end{equation}
identity at its null $c(\cdot)\equiv 0$.

\paragraph{Past-only derived path (covariate).} A governing law whose instantaneous residual is computable from the observed past, e.g.\ the conservation residual $\sum_{in}-\sum_{out}-dL/dt$, enters as an ordinary past-known covariate; it detects a present-state violation but does not anticipate onset.

\begin{figure}[h]
  \centering
  \includegraphics[width=0.85\textwidth]{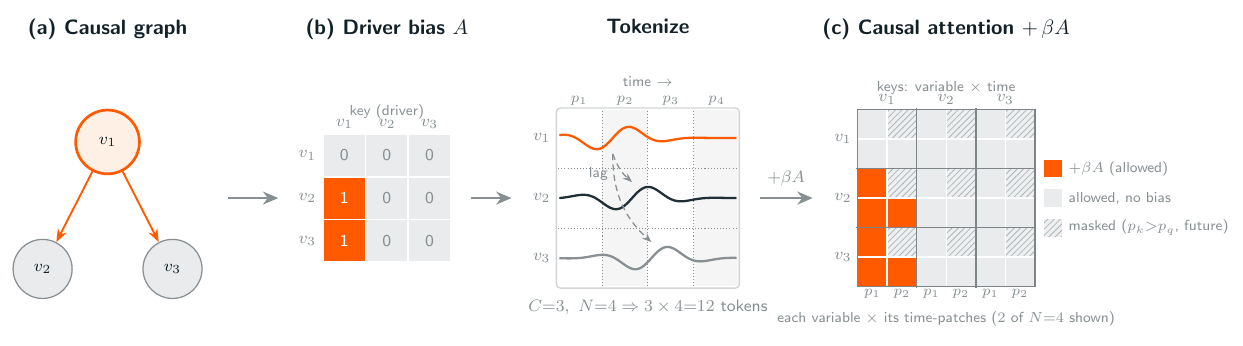}
  \caption{Causal-graph attention bias. A causal graph over variables ($v_2,v_3$ driven by $v_1$; a) gives the driver-adjacency $A$ (b). Since a token is a (variable, time-patch) pair, $A$ is tiled over all causally-allowed token pairs and added to the query--key logit as $\beta A$ (c; future keys masked). At $\beta{=}0$ the bias is the identity.}
  \label{fig:graph-bias}
\end{figure}

\paragraph{Attention bias.} A causal graph fixes a driver-adjacency $A\in\{0,1\}^{C\times C}$ ($A[a,b]{=}1$ iff $b$ drives $a$, zero diagonal), added to the attention logits,
\begin{equation}
  \mathrm{logit}(q_a, k_b) \mathrel{+}= \beta \cdot A[a,b] ,
  \label{eq:attention-bias}
\end{equation}
identity at $\beta=0$ (Figure~\ref{fig:graph-bias}, Appendix~\ref{app:beta}). One design point remains open: the predictor's future-known input is not yet cardinality-compatible across datasets with differing covariate counts (Appendix~\ref{app:covfig}).

\section{Dataset--prior alignment}
\label{app:alignment}

Table~\ref{tab:alignment} gives the full mapping between datasets and the semantic-prior categories they carry, referenced from Section~\ref{sec:negative}.

\begin{table}[h]
  \centering
  \caption{Dataset--prior alignment. Category numbers refer to Table~\ref{tab:taxonomy-ext}.}
  \label{tab:alignment}
  \footnotesize
  \begin{tabular}{@{}lll@{}}
    \toprule
    Dataset             & Prior categories                                    & Role                                       \\
    \midrule
    Occupancy           & Periodicity (\#5), event mechanism (\#12)           & periodicity clock (flagship)               \\
    BATADAL             & Conservation (\#10), control status (\#7)           & conservation residual                      \\
    ETTm1               & Identity (\#1), periodicity (\#5), smoothness (\#6) & static priors inert; full-context boundary \\
    Weather             & Units (\#2), noise (\#8), causal edges (\#9)        & negative control                           \\
    Beijing air-quality & Bounds/thresholds (\#3)                             & bounds prior                               \\
    SKAB                & Trend (\#4)                                         & trend prior                                \\
    GECCO               & Conservation (\#10)                                 & conservation-onset boundary                \\
    CMAPSS FD001        & Future-oracle covariate                             & non-deployable oracle ceiling              \\
    SYNTH-ROUTE-v2      & Structural identity (\#1)                           & synthetic structural prior                 \\
    \bottomrule
  \end{tabular}
\end{table}

\paragraph{Where priors do not help (full evidence).} The four causes summarised in Section~\ref{sec:negative} rest on the following runs. \emph{Information the window already carries}: static per-variable injections are inert across the static-injection ablations ($0/19$ survive BH-FDR; the real prior is indistinguishable from a shuffled prior even when isolated), consistent with a FiLM-style identity prior being provably absorbable into the following linear map; the periodicity clock is null once the context spans a full period (a full-context ETTm1 sweep: $0/3$ cells survive FDR); and the causal-graph structural prior is null once the bias is switched on at all ($375$ runs over five datasets with no positive cell in $19$; the earlier graph-correctness ablations are void, their arms having been byte-identical). \emph{A prior misaligned to the events}: a wrong FFT-estimated period reverses the Occupancy win to $-0.022$ (its phase precesses across the split), and the pre-registered Weather negative control is null ($-0.017$, $p=0.69$). \emph{Present state rather than anticipation}: the conservation residual detects an ongoing violation ($+0.741$ on the BATADAL ongoing stratum) but does not forecast onset before it begins (the onset-stratum runs: BATADAL onset $-0.053$; GECCO onset $-0.136$). \emph{Capacity rather than content}: the static-identity FiLM prior is absorbable by the next linear layer, and in the cross-domain transfer experiment gives only $+0.006$ ($p=0.56$, real $\approx$ shuffled).

\section{Cross-domain transfer: full results}
\label{app:transfer}

\begin{figure}[h]
  \centering
  \includegraphics[width=\textwidth]{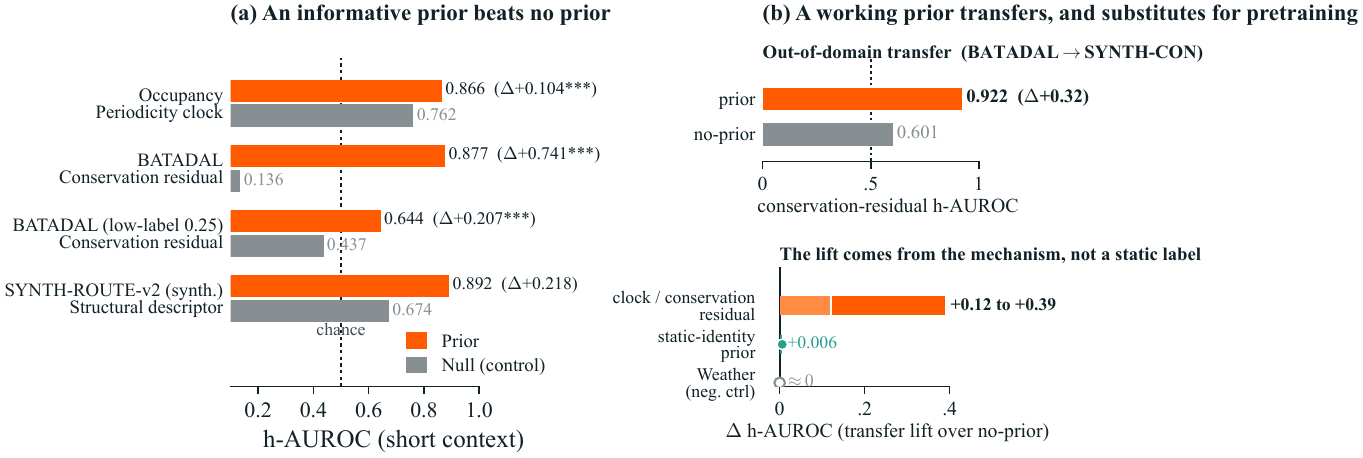}
  \caption{\textbf{An informative prior beats no prior} (h-AUROC; chance $0.5$, rank-based; each bar pairs the prior against the prior-free control of Table~\ref{tab:main}). \textbf{(a)}~In-domain, short context: each prior clears its null on the stratum where it applies (clock on silent-onset \emph{forecasting}; conservation residual on ongoing-violation \emph{detection} and at a $0.25$ label fraction; a structural descriptor on synthetic SYNTH-ROUTE-v2). \textbf{(b)}~Out-of-domain transfer, frozen encoder (BATADAL$\to$SYNTH-CON): a working prior transfers and \emph{substitutes} for feature transfer, while a static-identity prior does not.}
  \label{fig:results}
\end{figure}

This appendix expands the Claim-B result summarised in Section~\ref{sec:results}. Every run holds the target fixed and varies only where the encoder was pretrained (\texttt{id}, the target itself; \texttt{ood}, a source pool; \texttt{scratch}, no pretraining), with the prior delivered to the predictor at Stage 2; because the prior never touches the encoder, the arms in a cell are paired exactly off one cached Stage-1 checkpoint. Two pre-registered gates decide whether a cell is interpretable: G1 (mechanism-arrival) asks whether the prior helps in-domain at all, and G2 (transfer-substrate) asks whether source pretraining beats scratch. Both gates pass in exactly one cell, BATADAL$\to$SYNTH-CON under a frozen encoder, where the out-of-domain lift equals the in-domain lift (OOD$-$ID $=-0.002$ [$-0.024$, $+0.020$], $p=0.82$): a clean null for the amplification prediction.

The substitution reading is that a prior with no pretraining beats transfer with no prior in eight of ten cells, by $+0.027$ to $+0.381$, and is null exactly where the prior is known not to work (Weather, $+0.003$, $p=0.72$). The competence-gradient signal appears under the supervised source recipe on BATADAL$\to$SYNTH-CON, where the prior's lift rises as encoder competence falls (in-domain $+0.135$, out-of-domain $+0.195$, random encoder $+0.343$; \texttt{scratch}$-$\texttt{id} $=+0.208$ [$+0.040$, $+0.377$], $p=0.027$, with G2 passing at $p=0.05$), but that cell fails G1 by a ceiling effect (the in-domain model already scores $0.788$ without the prior), so the pre-registration marks it non-interpretable and we report it as suggestive only. The largest external-validity gap is that the primary conservation target is synthetic, because a second real mass-balance network (WADI, SWaT) was not licensable in the available time; the BATADAL source is real.

\section{Extended semantic-prior taxonomy}
\label{app:taxonomy}

Table~\ref{tab:taxonomy-ext} lists the twelve categories with an added implementation-status column.

\begin{table}[h]
  \centering
  \caption{Extended taxonomy with the finer per-category mechanism and implementation status. \textbf{E}: implemented and evaluated here; \textbf{I}: mechanism implemented; \textbf{N}: tested and found inert; \textbf{C}: conceptual scaffold. Static per-variable priors (\#1--\#4, \#6--\#8, \#12 via static injection) were found inert in the static-injection ablations; \#1 is evaluated via a structural per-variable descriptor on synthetic data.}
  \label{tab:taxonomy-ext}
  \footnotesize
  \setlength{\tabcolsep}{3.5pt}
  \resizebox{\textwidth}{!}{%
    \begin{tabular}{@{}clll c@{}}
      \toprule
      \# & Category                & Detailed mechanism                              & Example                                                        & Status \\
      \midrule
      1  & Identity                & Static covariate to encoder                     & High-voltage active load channel (ETTm1)                       & E      \\
      2  & Units/scale             & Static covariate to encoder                     & Temperature in ${}^\circ$C, $O(10)$ magnitude (Weather)        & N      \\
      3  & Bounds/thresholds       & Covariate; opt.\ past-only dist.-to-thr.        & Air-quality particulate warning $>250\,\mu g/m^3$ (BeijingAQ)  & C      \\
      4  & Trend                   & Static covariate to encoder                     & Thermocouple monotone, $\tau=+0.97$ (SKAB)                     & C      \\
      5  & Periodicity             & Future-known covariate to predictor; $P$ static & Daily load cycle $\to$ sin/cos, period $1440$ (ETTm1)          & E      \\
      6  & Smoothness              & Static covariate to encoder                     & Oil temperature slowly varying (ETTm1)                         & N      \\
      7  & Control/setpoint status & Static covariate to encoder                     & Pump actuator vs.\ passive sensor (BATADAL)                    & C      \\
      8  & Noise/intermittency     & Static covariate to encoder                     & Rainfall zero-inflated $\to$ intermittent (Weather)            & N      \\
      9  & Causal edges            & Attention bias ($\beta A$ driver-adjacency)     & Solar radiation drives temperature, $+0.54$ at $+3$h (Weather) & I      \\
      10 & Conservation            & Past-only residual covariate; law static        & Tank mass balance $\sum_{in}-\sum_{out}-dL/dt$ (BATADAL)       & E      \\
      11 & Position/distance       & Covariate or attention bias                     & Junction coordinates $(x,y)$; moving variable                  & C      \\
      12 & Event mechanism         & Static covariate to encoder                     & Occupancy onset: light leads, CO$_2$ lags                      & C      \\
      \bottomrule
    \end{tabular}%
  }
\end{table}

\section{Covariate mechanism illustration}
\label{app:covfig}

\FloatBarrier

\begin{figure}[h]
  \centering
  \includegraphics[width=0.9\textwidth]{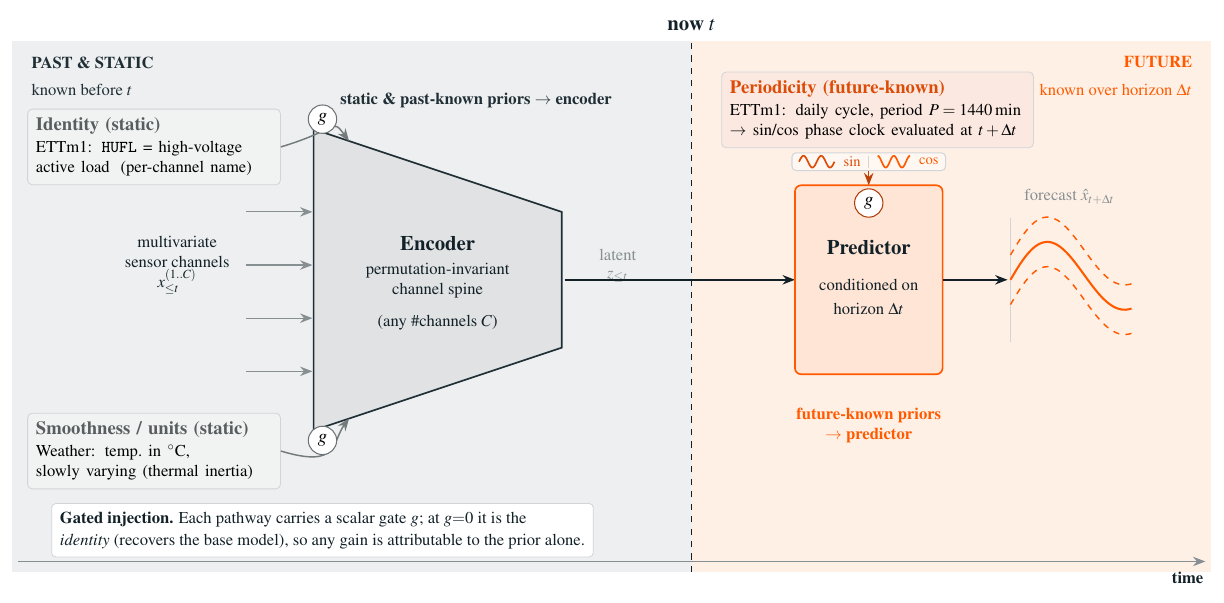}
  \caption{Covariate mechanism. Static and past-known priors (ETTm1 identity; Weather units and smoothness) enter the encoder through the zero-initialized identity-at-null pathway; the future-known periodicity clock ($P=1440$) enters the predictor at the forecast horizon.}
  \label{fig:covariate}
\end{figure}

Figure~\ref{fig:covariate} makes the routing decision of Section~\ref{sec:mechanisms} concrete on two datasets. ETTm1's channel identity (a static, per-channel fact: which channel is this transformer's high-voltage active load) and Weather's units and smoothness descriptors (both past-known per-channel facts) travel the static path of mechanism~(I). They reach the encoder through the zero-initialized identity-at-null connection of Equation~\ref{eq:gated-pathway}, so at initialization they contribute nothing and the encoder reduces exactly to the unconditioned baseline. The Occupancy periodicity clock ($P=1440$, Equation~\ref{eq:periodicity-clock}) takes the future-known path of mechanism~(I) instead. It skips the encoder entirely, is evaluated at the forecast horizon $\Delta t$, and is concatenated straight into the predictor's input, telling the model the future phase of the daily cycle rather than leaving it to infer that phase from a context window that may not span a full period.

What separates the two routes is what the prior actually conditions. A static or past-known descriptor shapes the encoder's representation of the observed window, while a future-known covariate shapes the predictor's forecast at a given horizon. Keeping them on separate pathways, rather than folding every covariate into one stream, is what lets Section~\ref{sec:negative} attribute a measured gain to the mechanism that carried it, static injection or future-known phase, instead of to a generic increase in input information.

\FloatBarrier

\section{Proofs}

\subsection{KITE tokenization}
\label{app:tokenization}

\begin{figure}[t]
  \centering
  \includegraphics[width=\textwidth]{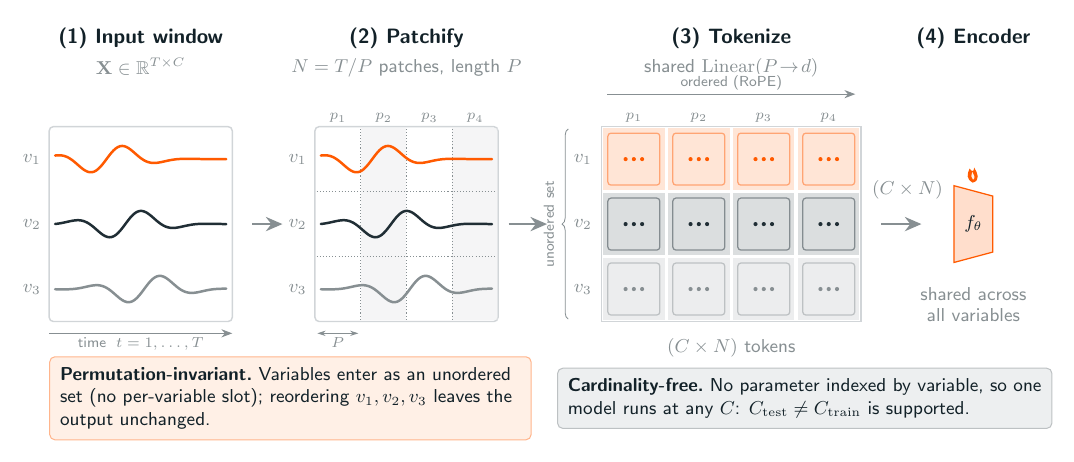}
  \caption{KITE tokenization. A $T\times C$ window is cut into $N=T/P$ temporal patches; a shared $\mathrm{Linear}(P\!\to\!d)$ maps each (variable, patch) to a token, yielding a $(C\times N)$ grid fed to the encoder $f_\theta$. Temporal patches are ordered (RoPE); variables form an unordered set with no per-variable parameter, making KITE permutation-invariant and cardinality-free ($C_{\text{test}}\neq C_{\text{train}}$).}
  \label{fig:tokenization}
\end{figure}

Figure~\ref{fig:tokenization} shows the tokenization procedure discussed in Section~\ref{sec:backbone}.

\subsection{Permutation invariance and cardinality freedom}
\label{app:perm}

We establish the two properties of the KITE backbone (Figure~\ref{fig:architecture}): the encoder output is unchanged when the variables are reordered, and the encoder works for any variable count $C$. Let $\pi$ be any reordering of variables and $x_{\le t}^\pi$ the input with variables reordered by $\pi$.

\paragraph{Embedding.} The window is cut into patches and every (variable, patch) is embedded by the same map $\mathrm{Linear}(P \to d)$: $z_{c,i} = W x_i^{(c)}$, where $W \in \mathbb{R}^{d\times P}$ is the shared embedding matrix, identical for all $(c,i)$. Because $W$ has no variable index, reordering the variables leaves the token set unchanged and only changes its order along the token axis, $Z(x_{\le t}^\pi) = \pi \cdot Z(x_{\le t})$.

\paragraph{Attention.} The encoder forms queries, keys, and values from tokens alone, and its causal mask restricts attention by temporal patch index, never by variable. Attention therefore depends only on token content, so reordering the input tokens produces the identically reordered output:
\begin{equation}
  \mathrm{Attn}(\pi \cdot Z) = \pi \cdot \mathrm{Attn}(Z).
  \label{eq:attn-equivariance}
\end{equation}

\paragraph{Pooling.} The final representation $\mathbf{h}_t$ is a symmetric aggregation over the token axis, and a symmetric function ignores the order of a set:
\begin{equation}
  \mathrm{agg}(\pi \cdot V) = \mathrm{agg}(V).
  \label{eq:agg-invariance}
\end{equation}

\paragraph{Composition.} Chaining the three stages,
\begin{equation}
  \mathbf{h}_t(x_{\le t}^\pi) = \mathrm{agg}\big(\mathrm{Attn}(\pi \cdot Z)\big) = \mathrm{agg}\big(\pi \cdot \mathrm{Attn}(Z)\big) = \mathrm{agg}\big(\mathrm{Attn}(Z)\big) = \mathbf{h}_t(x_{\le t}),
  \label{eq:permutation-invariance}
\end{equation}
so the reordering that enters at embedding disappears at pooling and $\mathbf{h}_t$ is independent of variable order. Every parameter is applied per token and none has a dimension equal to $C$: changing $C$ changes the number of token rows, not the shape of any parameter, so the trained $\theta$ runs at $C_{\text{test}} \ne C_{\text{train}}$ with nothing modified.

\subsection{\texorpdfstring{Identity at $\beta = 0$}{Identity at beta = 0}}
\label{app:beta}

The mechanism (Figure~\ref{fig:graph-bias}) adds the scaled driver-adjacency to the attention logits, $\mathrm{logit}(q_a, k_b) \mathrel{+}= \beta\, A[a,b]$, where $A \in \{0,1\}^{C\times C}$ is indexed by (query variable $a$, key variable $b$) and $A[a,b]=1$ iff $b$ is a causal driver of $a$ (a direct graph parent within its control loop) and $0$ otherwise, including the entire diagonal and every driver-less row. At $\beta=0$ the added term is $0\cdot A[a,b]=0$ for every pair, so the attention logits, and with them the whole forward pass, are identical to the unconditioned encoder at the same weights. As for the covariate paths, a trained comparison still needs a separately trained arm. The adjacency introduces no learned weights of its own, so the entire prior enters through the scalar $\beta$.

\section{Graph mechanism: correctness, correction, and the bounded null}
\label{app:graph}

The graph attention-bias mechanism assumes by default that the input causal graph is correct; here we characterise how an inaccurate graph behaves. The mechanism's only contact with the graph is the driver-adjacency matrix $A \in \{0,1\}^{C\times C}$, so a graph change that does not change $A$ does not change the model, and graph correctness is therefore defined on $A$, equivalently on the set of (target, driver) pairs it flags. Let $A^*$ be the driver-adjacency of the ground-truth graph and $\hat A$ that of the graph under test, with flagged pair sets $E^* = \{(a,b): A^*[a,b]=1\}$ and $\hat E = \{(a,b): \hat A[a,b]=1\}$. Graph correctness is the overlap
\begin{equation}
  \gamma = \frac{|E^* \cap \hat E|}{|E^* \cup \hat E|} \in [0,1],
  \label{eq:jaccard}
\end{equation}
where $\gamma=1$ flags exactly the right driver pairs and $\gamma=0$ flags none correctly. The Jaccard form ignores the many $C^2$ entries both graphs correctly leave unflagged, which would otherwise make almost any graph look correct, and scores only the pairs the bias acts on.

\paragraph{The corrected null and its bound.} An early ablation compared graphs at the released $\beta=0$, where the additive bias vanishes identically, so correct, shuffled, random, and dense adjacencies produce byte-identical activations; those arms are therefore void. Fixing $\beta=2$ switches the mechanism on for the first time, and it demonstrably acts: the share of cross-variable attention mass landing on graph edges rises from a $3.8\%$ chance level to $19.1\%$ on Weather and from $1.6\%$ to $8.7\%$ on the Tennessee Eastman process (TEP)~\citep{downs_tep_1993,rieth_tep_2017}. Even so it does not help. Across seven adjacency variants on Weather every paired difference in h-AUROC falls within $\pm0.0013$, against a seed standard deviation of $0.0038$, and closing every cross-variable path inside the encoder costs only $-0.0011$ h-AUROC ($95\%$ CI $[-0.0063,+0.0040]$, $n=5$), bounding what any variable-relational prior could carry here at about $0.006$. This is a bound, not a refutation, because the positive control is under-powered: on a synthetic benchmark whose event shifts two variables' joint distribution while both marginals stay fixed, a linear probe handed all $45$ pairwise correlations reaches only $0.559$ against a variable-blind floor of $0.454$, with KITE at $0.478$--$0.519$ inside that corridor. We therefore report mechanism~(II) as undetermined with a small upper bound rather than refuted.

\section{Implementation alignment}
\label{app:implalign}

The results in Section~\ref{sec:results} were produced by an implementation that differed from the
design described in Section~\ref{sec:backbone} and Section~\ref{sec:mechanisms} in four places. A
companion study re-ran both flagships under the idealised form of each, separately and together, on
the same data, splits, seeds and metric, changing one thing at a time. Every arm below is paired by
seed against its own identity-at-null control (the covariate present but zeroed), $10$ seeds on
Occupancy and $5$ on BATADAL, with Benjamini--Hochberg control within each flagship.

\paragraph{Two changes have no detectable cost.} Removing the per-variable slot and supplying the
future-known covariate during Stage-1 pretraining each leave every flagship without a detectable
change (all $p \ge 0.08$, Table~\ref{tab:implalign}); at these sample sizes this is a failure to
reject, not demonstrated equivalence. This paper describes the slot-free encoder; the Stage-1
covariate variant is not adopted, for the reason given below.

\paragraph{One costs accuracy, and one gives no reason to switch.} A scalar-gated form of the
fusion in Equation~\eqref{eq:gated-pathway}, $\mathrm{out}=x+\tanh(g)\,\delta$, costs accuracy on
every flagship (Table~\ref{tab:implalign}), so this paper describes the zero-initialized fusion.
Substituting SIGReg for VICReg is neither better nor detectably worse on any flagship (below), so
the VICReg collapse term is kept.

\paragraph{A distinction the study had to make.} Equation~\eqref{eq:gated-pathway}'s
``present during pretraining'' refers to the \emph{static descriptor entering the encoder}, and that
is what the implementation does: the descriptor is concatenated to the Stage-1 input alongside the
observed variables, so the encoder is trained with it and is frozen afterwards with it. The
\emph{future-known} covariate of the predictor path is a separate mechanism and is deliberately not present in Stage 1: the predictor is retrained at Stage 2 anyway, and
keeping the covariate out of pretraining keeps the encoder's Stage-1 data identical in the prior
and no-prior arms. The row below measures the variant that supplies it during pretraining and
finds no detectable cost, so nothing is lost by leaving it out.

\begin{table}[h]
  \centering
  \caption{Prior-versus-null $\Delta$ h-AUROC under each idealised change, against the identity-at-null (zeroed-covariate) control. The \emph{shipped} row is this study's paired re-run of the configuration behind Table~\ref{tab:main}, reproducing each flagship effect up to seed noise. \textbf{Cost} is the paired per-seed difference in the prior arm's h-AUROC from that shipped configuration (so it need not equal the change in $\Delta$, whose null arm can also move).}
  \label{tab:implalign}
  \small
  \resizebox{\textwidth}{!}{%
    \begin{tabular}{@{}llccc@{}}
      \toprule
      Configuration                              & Change   & Occupancy (silent-onset) & BATADAL (ongoing)    & BATADAL ($0.25$ labels) \\
      \midrule
      shipped (re-run of Table~\ref{tab:main})   & $\Delta$ & $+0.096$                 & $+0.736$             & $+0.205$                \\
      no per-variable slot                       & $\Delta$ & $+0.092$                 & $+0.510$             & $+0.192$                \\
                                                 & cost     & $-0.011$ ($p=0.158$)     & $-0.031$ ($p=0.537$) & $-0.007$ ($p=0.664$)    \\
      scalar-gated fusion, $\tanh(g)$            & $\Delta$ & $+0.041$                 & $+0.166$             & $+0.099$                \\
                                                 & cost     & $-0.055$ ($p=0.015$)     & $-0.570$ ($p<0.001$) & $-0.106$ ($p=0.061$)    \\
      future-known covariate also at pretraining & $\Delta$ & $+0.089$                 & $+0.745$             & $+0.195$                \\
                                                 & cost     & $-0.006$ ($p=0.551$)     & $+0.009$ ($p=0.713$) & $-0.010$ ($p=0.084$)    \\
      all three together                         & $\Delta$ & $+0.037$                 & $+0.227$             & $+0.116$                \\
                                                 & cost     & $-0.066$ ($p=0.004$)     & $-0.314$ ($p=0.104$) & $-0.083$ ($p=0.159$)    \\
      \bottomrule
    \end{tabular}%
  }
\end{table}

\paragraph{The gated fusion is the expensive change.} Implemented faithfully, $g$ ends training at a
magnitude of $0.029$ on Occupancy and $0.007$ on BATADAL: the pathway is gated rather than dead ($g$
is never exactly zero) but it admits only a few percent of the descriptor. The mechanism is
visible across and within flagships: the flagship whose gate opens less loses proportionally more
($77\%$ against $53\%$), and within each flagship the seeds whose gate opened wider score higher
(Pearson $r=+0.89$ and $+0.81$, $n=10$; correlational across seeds, so supporting evidence rather
than a causal demonstration). Nothing in the objective rewards opening the gate, because at $g=0$
the gradient reaching $g$ is proportional to how useful the descriptor would be if it were already
admitted. The zero-initialized fusion avoids this: the descriptor enters at full scale from the
first step and only its weights are learned.

\paragraph{SIGReg is not adopted.} Against the otherwise-identical configuration, neither
SIGReg~\citep{sigreg} nor a Sliced-Wasserstein variant is better on any flagship (all $p \ge 0.08$),
so the pre-registered adoption rule (non-inferior everywhere and strictly better somewhere)
returns \emph{keep VICReg}. The collapse diagnostics nonetheless move: both sketching objectives
raise the per-feature standard deviation from $0.089$ to $0.52$--$0.64$ against VICReg's hinge target
of $1$, and the Sliced-Wasserstein variant additionally reduces mean pairwise cosine from $0.956$ to
$0.734$, which the Epps--Pulley statistic cannot do because centering each one-dimensional sketch
makes it invariant to a shared mean direction. Representation quality and task score are therefore
decoupled on these benchmarks.

\paragraph{Permutation invariance and cardinality freedom, measured.} With the slot removed, the
encoder summary is invariant to variable reordering to the floating-point rounding floor
($\max_\pi \lVert \mathbf{h}_t(x^\pi) - \mathbf{h}_t(x) \rVert_\infty \approx 2\times10^{-7}$ over
$20$ permutations, both readouts, with and without a padding mask); with the slot present the same
quantity is $10^{-2}$ to $2\times10^{-1}$. On a \emph{trained} model, however, the downstream cost of
that violation is negligible: over $50$ permutations of the test set the slot-free model moves by
exactly $0$ and the slot model by at most $8\times10^{-4}$ h-AUROC, an order of magnitude below the
$0.009$ run-to-run floor this pipeline produces from initialization alone. One slot-free encoder also
runs unmodified at every variable count tested ($C_{\text{test}} \in \{2,\dots,10\}$ against
$C_{\text{train}}=5$, h-AUROC $0.822$ to $0.908$), whereas the slot encoder cannot be evaluated at any
$C \ne C_{\text{train}}$ at all.

\paragraph{A bound on the BATADAL row.} Scored through the same evaluation code, on the same windows,
strata, and re-run as this study's shipped row, a zero-parameter threshold on the conservation residual
itself reaches $0.921$ on the ongoing stratum, above the model's own $0.873$, and $0.489$ (chance) on
onset. That row
is therefore a detection result on which a training-free detector suffices, consistent with the
deterministic-detector column of Table~\ref{tab:main}, and it carries no evidence about anticipation
in either direction.

\section{Prior-only and generic-routing baselines}
\label{app:prioronly}

A prior-versus-no-prior contrast holds the architecture fixed. Table~\ref{tab:prioronly} asks the
converse: how far does the prior get on its own, and does it matter where it enters? Every scorer is
evaluated on the same test windows and stratum as the corresponding KITE result.

\begin{table}[h]
  \centering
  \caption{Prior-only and generic-routing baselines (h-AUROC). Prior-only scorers use the prior and
  nothing else: no sensors, no encoder, no pretraining. ``Input channels'' feeds the clock's past
  phase to the encoder as extra variables, the way any covariate-capable forecaster would take it.
  Occupancy KITE arms are means over $10$ seeds, BATADAL over $5$.}
  \label{tab:prioronly}
  \footnotesize
  \begin{tabular}{@{}lll c@{}}
    \toprule
    Row (estimand)                           & Scorer                                      & Uses              & h-AUROC   \\
    \midrule
    Occupancy (silent-onset)                 & no prior (KITE backbone)                    & sensors           & $0.762$   \\
                                             & clock only, time-of-day sine                & prior             & $0.721$   \\
                                             & clock only, logistic fit on the calendar    & prior             & $0.078$   \\
                                             & KITE, clock as encoder input channels       & sensors + prior   & $0.797$   \\
                                             & KITE, clock to the predictor at the horizon & sensors + prior   & $\mathbf{0.866}$ \\
    \midrule
    BATADAL (ongoing)                        & no prior (zeroed)                           & sensors           & $0.136$   \\
                                             & residual only, threshold (0 parameters)     & prior             & $\mathbf{0.921}$ \\
                                             & residual only, logistic regression          & prior             & $0.816$   \\
                                             & KITE, residual as covariate                 & sensors + prior   & $0.877$   \\
    \midrule
    BATADAL (overall, full labels)           & residual only, threshold                    & prior             & $0.631$   \\
                                             & KITE, residual as covariate                 & sensors + prior   & $0.637$   \\
    \bottomrule
  \end{tabular}
\end{table}

The two flagships answer differently. On Occupancy the clock alone is weaker than no prior at all,
and a fitted calendar model inverts because the finetuning split spans under two days, so the
calendar's phase does not carry across periods. What converts phase into a hazard is the
architecture, and where the clock enters matters: as ordinary input channels it adds $+0.035$, at
the horizon $+0.104$. On BATADAL the residual alone is sufficient: a threshold on it outranks the
trained pipeline on the ongoing stratum and matches it overall. That row therefore shows the value
of the residual as a detection feature, not of the architecture, as Appendix~\ref{app:implalign}
notes. We did not run the external forecasters of Table~\ref{tab:main} with the prior as an extra
input; the encoder-input arm above is the within-backbone version of that baseline.

\section{Reproducibility}
\label{app:repro}

All experiments use public benchmarks. The real datasets are Occupancy Detection~\citep{candanedo_occupancy_2016,candanedo_occupancy_uci_2016}, the BATADAL water-distribution challenge data~\citep{taormina_batadal_2018}, ETTm1~\citep{zhou_informer_2021}, the Beijing multi-site air-quality data~\citep{zhang_beijing_2017,chen_beijing_uci_2017}, SKAB~\citep{katser_skab_2020}, GECCO water-quality~\citep{moritz_gecco_2018}, the Weather benchmark (the Jena weather-station record, \url{https://www.bgc-jena.mpg.de/wetter/}, as released with~\citep{wu_autoformer_2021}), CMAPSS FD001~\citep{saxena_damage_2008}, and the Tennessee Eastman process~\citep{downs_tep_1993,rieth_tep_2017}, all obtained from their standard public releases (the UCI Machine Learning Repository for Occupancy and Beijing air-quality; the BATADAL challenge, GECCO challenge, and SKAB distributions; the long-horizon forecasting benchmark collection for ETTm1 and Weather; the NASA prognostics repository for CMAPSS; and Harvard Dataverse for TEP). The synthetic datasets (SYNTH-ROUTE-v2 and SYNTH-CON) are generated by the released code from fixed seeds; their generators, together with the typed semantic priors and per-dataset causal graphs, will be released with the code. Typed priors are provided for BATADAL, Beijing air-quality, ETTm1, GECCO, SKAB, and Weather, and causal graphs for all $22$ datasets.

The primary metric is h-AUROC, the AUROC of the event-probability surface pooled and averaged over forecast horizons $1,\dots,K$ (Section~\ref{sec:backbone}). Every prior-versus-null comparison is paired by seed under an otherwise identical recipe, so the only intended difference between arms is the prior input; each arm is pretrained and finetuned independently, except in the transfer study (Appendix~\ref{app:transfer}), where the prior is delivered predictor-side and the arms share one cached Stage-1 checkpoint. In-domain effects are reported over $5$ to $10$ seeds and transfer effects over $5$ seeds, with a zeroed-prior and a shuffled-prior control ladder; significance uses paired tests with Benjamini--Hochberg FDR control at $0.05$ across the multiple-comparison families noted in each result. The external baselines (Chronos-2, and a PatchTST/iTransformer backbone) are run over $5$ seeds on the same splits and strata, with the deterministic-detector arms run once. The complete configuration files, checkpoints manifest, and per-seed result tables will be released together with the code.

\end{document}